\documentclass[10pt,twocolumn,letterpaper]{article}

\usepackage{cvpr}
\usepackage{times}
\usepackage{epsfig}
\usepackage{graphicx}
\usepackage{amsmath}
\usepackage{amssymb}
\usepackage{algorithmic}
\usepackage[linesnumbered,ruled,vlined]{algorithm2e}
\usepackage{helvet}   
\usepackage{courier}  
\usepackage{url}      
\usepackage{xcolor}
\usepackage{booktabs}
\usepackage{multirow}
\usepackage{subcaption}
\usepackage[export]{adjustbox}

\DeclareMathAlphabet      {\mathbfit}{OML}{cmm}{b}{it}

\usepackage{pifont}

\renewcommand{\etal}{\textit{et~al}\mbox{.}}

\renewcommand{\ie}{i.e.,\ }
\newcommand{\bbR}{{\mathbb{R}}}

\usepackage[breaklinks=true,bookmarks=false]{hyperref}

\cvprfinalcopy 

\def\cvprPaperID{477} 

\begin{document}

\title{Learning Shape Representations for Clothing Variations\\ in Person Re-Identification}
\author{
Yu-Jhe~Li
\hspace{10.0mm}
Zhengyi Luo
\hspace{10.0mm}
Xinshuo Weng
\hspace{10.0mm}
Kris M. Kitani
\\
The Robotics Institute, Carnegie Mellon University
\\
{\tt\small \{\url{yujheli}, \url{zluo2}, \url{xinshuow}, \url{kmkitani}\}\url{@andrew.cmu.edu}}
}

\twocolumn[{%
\renewcommand\twocolumn[1][]{#1}%
\maketitle
\vspace{-14mm}
\begin{center}
    \centering
    \includegraphics[width=0.9\textwidth]{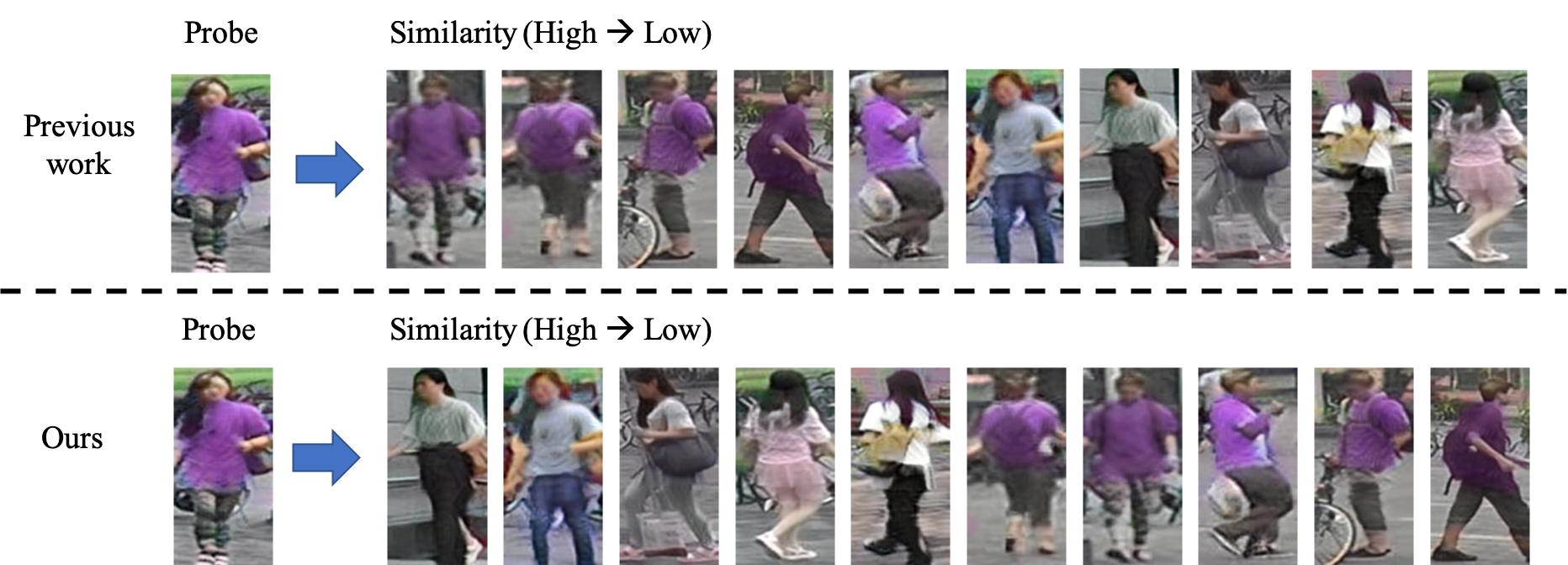}
    \captionof{figure}{Given a probe image and ten gallery figures from our synthesized dataset (five from same identity and five with same color of clothes as the probe), we aim to prioritize matching the images with same body shape though in different wearings while previous work~\cite{sun2018beyond,zheng2019joint} are dominated by clothing color information.}
    \label{fig:teasor}
\end{center}%
}]


\begin{abstract}
Person re-identification (re-ID) aims to recognize instances of the same person contained in multiple images taken across different cameras. Existing methods for re-ID tend to rely heavily on the assumption that both query and gallery images of the same person have the same clothing. Unfortunately, this assumption may not hold for datasets captured over long periods of time (e.g., weeks, months or years). To tackle the re-ID problem in the context of clothing changes, we propose a novel representation learning model which is able to generate a body shape feature representation without being affected by clothing color or patterns. We call our model the Color Agnostic Shape Extraction Network (CASE-Net). CASE-Net learns a representation of identity that depends only on body shape via adversarial learning and feature disentanglement. Due to the lack of large-scale re-ID datasets which contain clothing changes for the same person, we propose two synthetic datasets for evaluation. We create a rendered dataset SMPL-reID with different clothes patterns and a synthesized dataset Div-Market with different clothing color to simulate two types of clothing changes. The quantitative and qualitative results across 5 datasets (SMPL-reID, Div-Market, two benchmark re-ID datasets, a cross-modality re-ID dataset) confirm the robustness and superiority of our approach against several state-of-the-art approaches.
\end{abstract}

\begin{figure*}[t]
  \centering
  \begin{subfigure}[b]{0.45\linewidth}
    \centering\includegraphics[width=\linewidth]{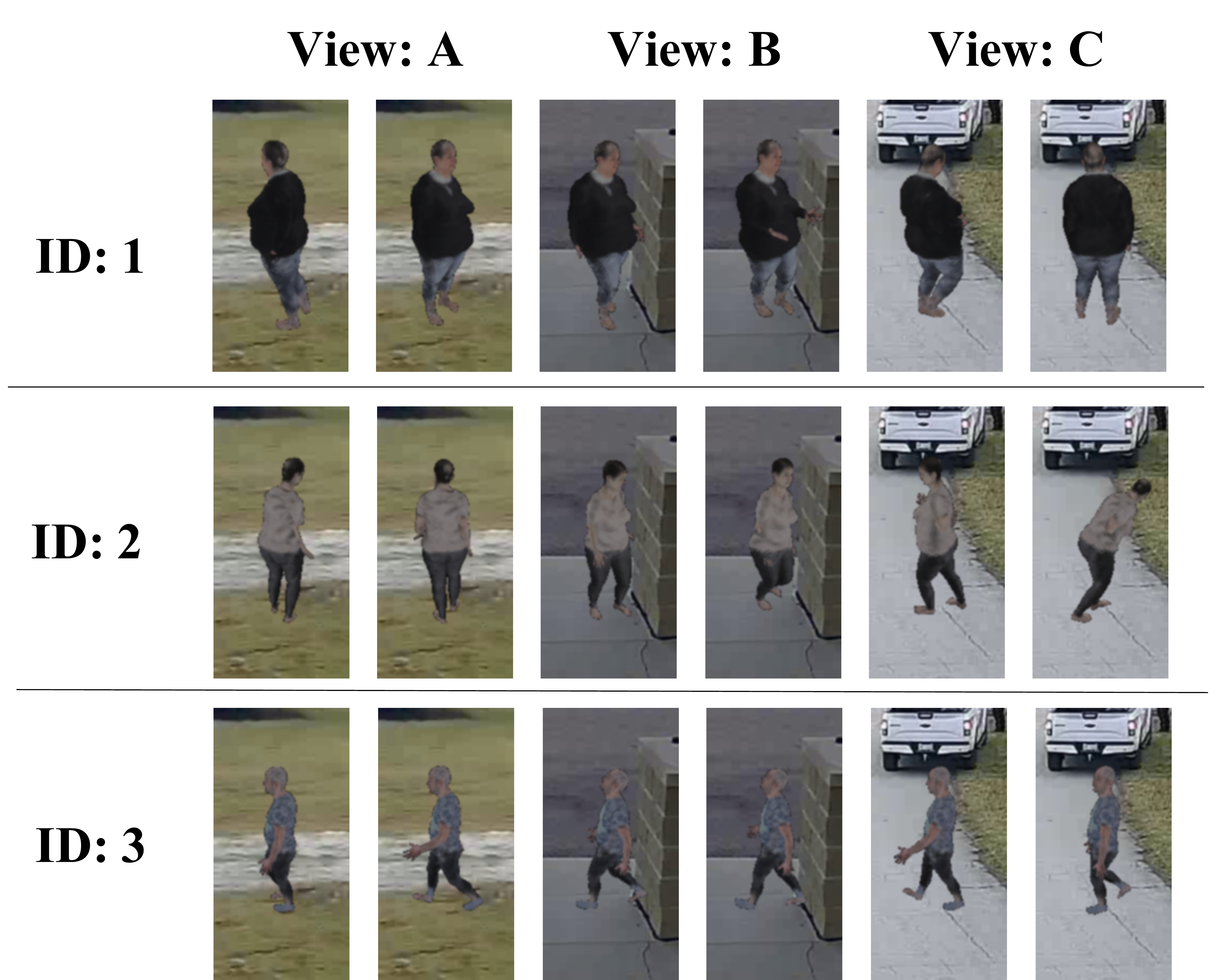}\\
    \caption{Training set (w/o clothes changes).}
    \label{fig:smpl-trn}
  \end{subfigure}
  \begin{subfigure}[b]{0.45\linewidth}
    \centering\includegraphics[width=\linewidth]{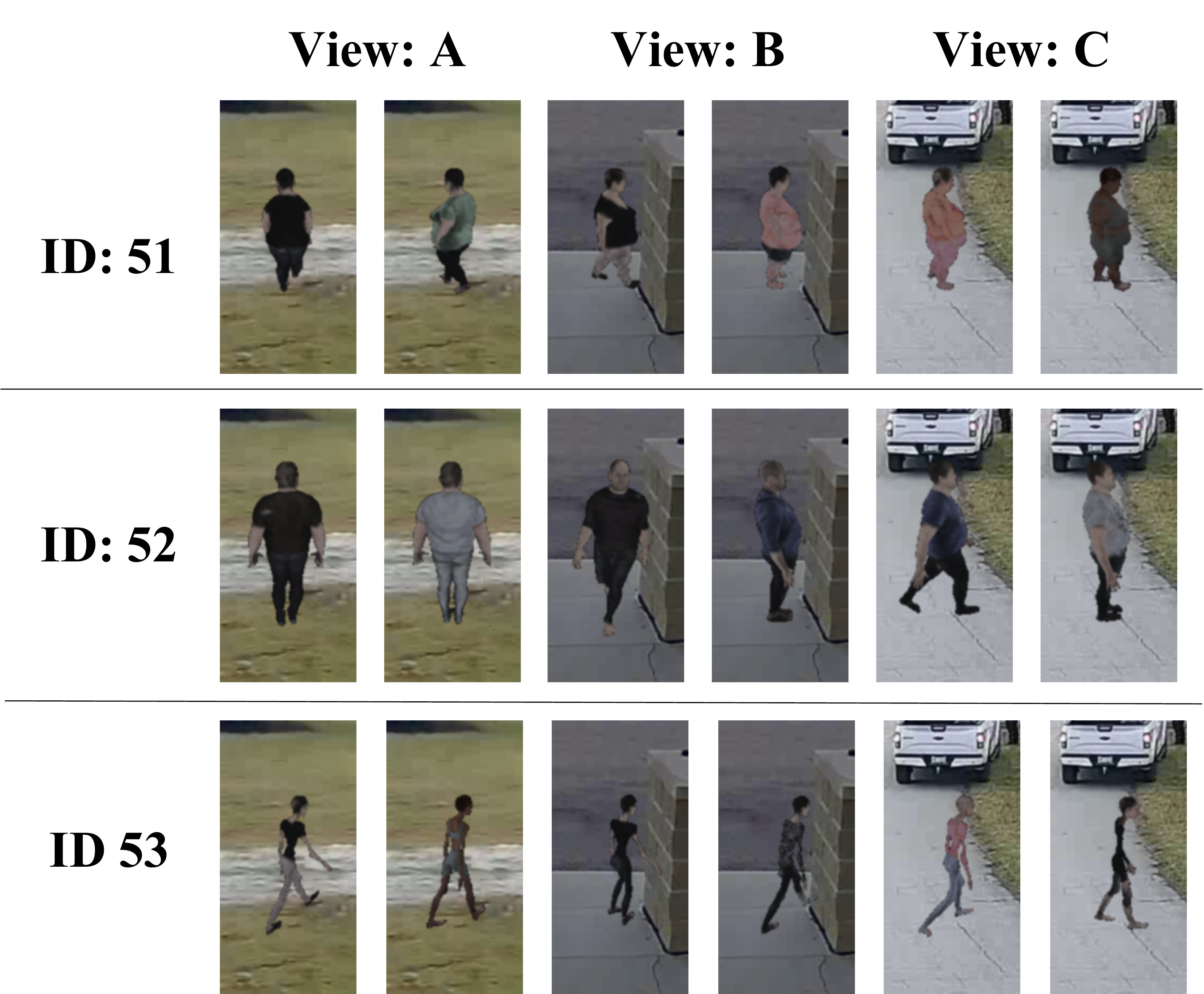}\\
    \caption{Testing set (w/ clothes changes).}
    \label{fig:smpl-tst}
  \end{subfigure}
  \vspace{-2.0mm}
  \caption{\textbf{Examples of our rendered re-ID dataset \emph{SMPL-reID}.} (a) The training set different identities with pose variations across different viewpoints, each of which is displayed in each row as example. We note that, no change in clothes occur in the training set.
  (b) The testing set contains the other identities with changes in clothes across different viewpoints.
  }
  \label{fig:smpl}
\end{figure*}

\begin{figure}[t]
  \centering
  \includegraphics[width=0.9\linewidth]{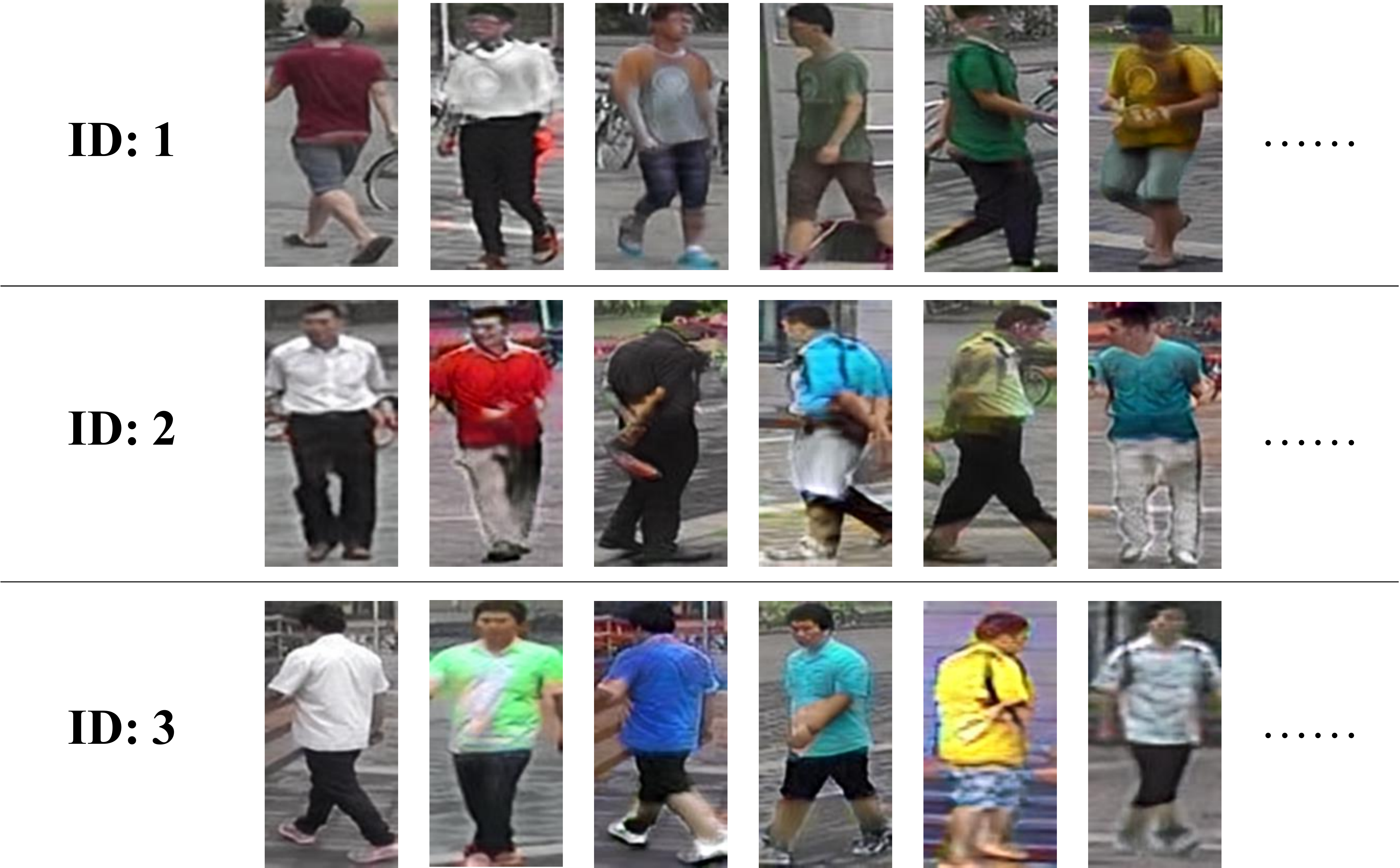}
  \caption{\textbf{Examples of our synthesized testing dataset \emph{Div-Market}.} Div-Market is synthesized from Market-1501 with changes in clothing color. The example of each identity is shown in each row.}
  \vspace{-2.0mm}
  \label{fig:div-market}
\end{figure}
Person re-identification (re-ID)~\cite{zheng2016person, Wu2016} aims to recognize the same person across multiple images taken by different cameras or different times. Re-ID is an important component technology in many applications such as person tracking~\cite{andriluka2008people}, video surveillance system~\cite{khan2016person} and computational forensics~\cite{vezzani2013people}. Despite promising performance, existing methods~\cite{hermans2017defense,lin2017improving,zhong2017camera,si2018dual,chen2018group,shen2018deep} rely (usually inadvertently) on the assumption that both query and gallery images of the same person will have the \emph{same clothing}. While this assumption does hold for many of the existing re-ID datasets, the same clothing assumption will not hold if the data has been aggregated over a long time span, since people tend to change their clothes daily.  For many re-ID methods, changing the clothing of an individual across images will result in a severe drop in performance.

Since popular re-ID datasets (\emph{e.g.}, Market1501~\cite{zheng2015scalable} and DukeMTMC-reID~\cite{zheng2017unlabeled,ristani2016MTMC}) lack clothing changes for the same identity, the \textit{clothing-dependence problem} has received little attention in previous work. Yet, Zheng~\etal~\cite{zheng2019joint} proposed a joint learning framework named DG-Net that involves a generative module that separately encodes each person into an appearance representation and a structure representation, which leads to improved re-ID performance in the existing re-ID datasets (\emph{e.g.}, Market-1501, DukeMTMC-reID). Although their idea of disentangling structure (shape) and appearance (color) should be useful for addressing the clothing-dependence problem, we show later that their use of appearance to perform re-ID is still dominated by clothing color information. As depicted in Fig.~\ref{fig:teasor}, given one probe image and ten gallery images (five from same identity and five with same clothing color as the probe), existing methods~\cite{sun2018beyond,zheng2019joint} are trained to focus only on matching clothing color while ignoring other identity-related cues such as the body shape.  

One could make the argument that methods for cross-modality re-ID ~\cite{wu2017rgb,ye2018hierarchical,ye2018visible,dai2018cross,wang2019rgb} address a very similar problem, which is the sensor-dependence problem. Sensor-dependence is indeed similar to the clothing-dependence problem as the re-ID system learns to leverage dataset bias to perform identification. Cross-modal re-ID methods force the model to discover sensor-invariant features by ensuring that the learned representation is agnostic to multiple input modalities (\emph{e.g.}, RGB images and IR images). At a high-level we will take a similar approach by forcing our representation to be invariant to clothing color and patterns by ensuring that the learned representation works for both color and gray scale images. However, we go further and explicitly guide the representation learning towards learning shape information.

To overcome the limitations of previous work and to confirm the clothing-dependence problem of modern re-ID methods, collecting a new dataset containing changes in clothes is an option. However, this can cause ethical issues with privacy and several existing datasets such as DukeMTMC was suspended as a result. In order to address the clothing-dependence problem of modern re-ID methods we need a dataset on which to evaluate. We create a synthetic dataset named \emph{SMPL-reID} to simulate clothing changes for the same person using computer graphics. We render $100$ identity across $6$ view-points (different camera angles) and $10$ walking poses using SMPL\cite{Loper2015}, which is a parametric human body mesh model. As shown in Fig.~\ref{fig:smpl}, $50$ identities are used for training where no clothing change occurs, and the remaining $50$ identities in the testing dataset contain changes in clothes. This will allow us to evaluate whether a re-ID method can really handle changes in clothing at test time. We also synthesize another dataset \emph{Div-Market} with diverse changes in clothing \textit{color} (no change of clothing type) across images of the same identity from Market1501~\cite{zheng2015scalable} using a generative model (instead of computer graphics rendering). As depicted in Fig.~\ref{fig:div-market}, we synthetically change the color or texture of the original clothing. In general, these two rendered and synthesized dataset are developed to evaluate the weakness of existing state-of-the-art approaches in the scenario of changing clothes. We confirm that all of our compared re-ID approaches exhibit severe performance drop in the experiments.



In addition to the aforementioned datasets, we propose a strong baseline framework \emph{Color Agnostic Shape Extraction Network (CASE-Net)} to address the clothing-dependence problem, which learns body-structural visual representations via adversarial learning and structural disentanglement. In particular, we leverage gray-scaled images produced from the RGB ones to derive visual features of same distribution across clothing color variations. Moreover, our model achieves structural distilling by performing image recovery (generation) when observing both gray-scaled and RGB images with pose variations. Our experiments show that, compared to prior re-ID approaches which are susceptible to errors when clothing changes, our model is able to achieve state-of-the-art performance on the synthesized clothing-color changing datasets. To the best of our knowledge, we are among the first to address the issue with clothing change for re-ID. The contributions of this paper are highlighted below:
\begin{itemize} 
  \item We collect two re-ID datsets (\emph{SMPL-reID} and \emph{Div-Market}) for simulating real-world scenario, which contain changes in clothes or clothing-color for evaluating the weakness of existing re-ID approaches.
  \item We also propose an end-to-end trainable network which advances adversarial learning strategies across color variations for deriving body shape features and image generation across pose variations to achieve body shape disentanglement.
  \item The quantitative and qualitative results on 5 datasets (our collected \emph{SMPL-reID}, \emph{Div-Market}, two standard benchmark datasets, one cross-modality dataset) confirm the weakness of the current state-of-the-art approaches and the effectiveness as well as the generalization of our method.
\end{itemize}




\section{Related Works} \label{sec:related}


\paragraph{\textbf{Person Re-ID.}}
Person re-ID has been widely studied in the literature. Existing methods typically focus on tackling the challenges of matching images with viewpoint and pose variations, or those with background clutter or occlusion presented~\cite{li2019greid,cheng2016person,lin2017improving,kalayeh2018human,si2018dual,chang2018multi,li2018harmonious,liu2018pose,wei2018person,song2018mask,chen2018group,shen2018deep,wang2018resource,sun2018beyond,chen2019learning,li2018adaptation}. For example, Liu~\etal~\cite{liu2018pose} develop a pose-transferable deep learning framework based on GAN~\cite{goodfellow2014generative} to handle image pose variants. Chen~\etal~\cite{chen2018group} integrate conditional random fields (CRF) and deep neural networks with multi-scale similarity metrics. Several attention-based methods~\cite{si2018dual,li2018harmonious,song2018mask} are further proposed to focus on learning the discriminative image features to mitigate the effect of background clutter. While promising results have been observed, the above approaches cannot easily be applied for addressing clothing dependence problem due to the lack of ability in suppressing the visual differences across clothes/clothing-colors.
\paragraph{\textbf{Cross-modality Re-ID.}}
Some related methods for cross-modality are proposed~\cite{wu2017rgb,ye2018hierarchical,ye2018visible,dai2018cross,wang2019rgb,hao2019hsme} to address color-variations.
Varior~\etal~\cite{varior2016learning} learned color patterns from pixels sampled from images across camera views, addressing the challenge of different illuminations of the perceived color of subjects. Wu~\etal~\cite{wu2017rgb} built the first cross-modality RGB-IR benchmark dataset named SYSU-MM01. They also analyze three different network structures and propose deep zero-padding
for evolving domain-specific structure automatically in one stream network optimized for RGB-IR Re-ID tasks. \cite{ye2018hierarchical,ye2018visible} propose modality-specific and modality-shared metric losses and a new bi-directional dual-constrained top-ranking loss for RGB-Thermal person re-identification. \cite{dai2018cross} introduce a cross-modality generative adversarial network (cmGAN) to reduce the distribution divergence of RGB and IR features.  Recently, \cite{wang2019rgb} also propose pixel alignment and feature alignment jointly to reduce the cross-modality variations. Yet, even though they successfully learn sensor-invariant features across modality, their models still can not be used to address the issue of clothing dependence in single modality. 

\paragraph{\textbf{Disentanglement Re-ID.}}
Recently, a number of models are proposed to better represent specific disentangled features during re-ID~\cite{su2017pose,zheng2017pose,zhao2017deeply,zhao2017spindle,li2017learning,yao2019deep,wei2017glad}. Ma~\textit{et al.}~\cite{ma2018disentangled} generate person images by disentangling the input into foreground, background and pose with a complex multi-branch model which is not end-to-end trainable. Ge~\textit{et al.}~\cite{ge2018fd} and Li~\etal~\cite{li2019cross} learn pose-invariant features with guided image information. Zheng~\etal~\cite{zheng2019joint} propose a joint learning framework named DG-Net that couples re-id learning and data generation end-to-end. Their
model involves a generative module that separately encodes
each person into an appearance code and a structure code,
which lead to improved re-ID performance. However, their appearance encoder used to perform re-ID are still dominated by the clothing-color features corresponding to the input images. Based on the above observations, we choose to learn clothing-color invariant features using a novel and unified model. By disentangling the body shape representation, re-ID can be successfully performed in the scenario of clothing change even if no ground true images containing clothing change are available for training data.
\begin{figure*}[t]
  \centering
  \includegraphics[width=0.9\linewidth]{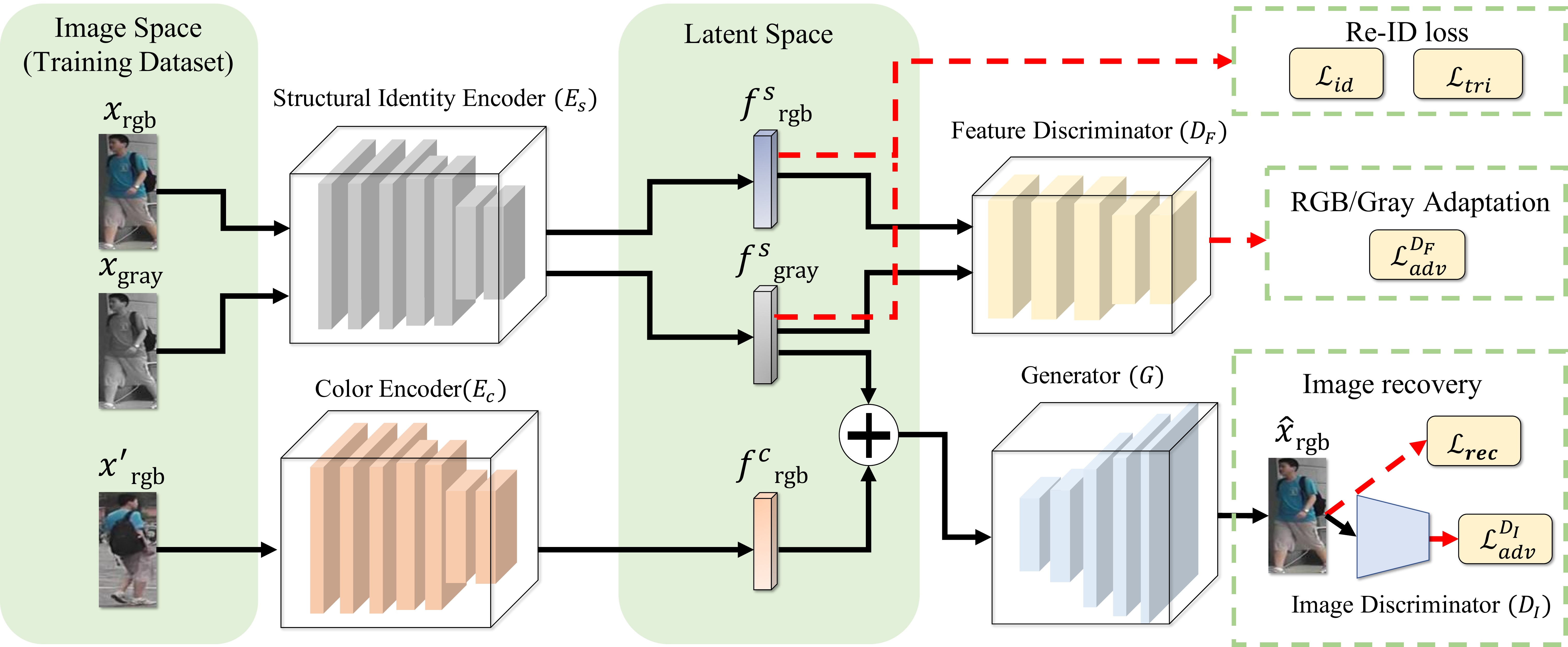}
  \vspace{-2mm}
  \caption{\textbf{Overview of the proposed Color Agnostic Shape Extraction Network (CASE-Net)}. The shape encoder $E_S$ encodes input images across different color domains/datasets ($x_{\textrm{rgb}}$ and $x_{\textrm{gray}}$) and produces color-invariant features $f^s$ ($f^s_{\textrm{rgb}}$ and $f^s_{\textrm{gray}}$). The color encoder $E_C$ encodes the RGB images ($x_{\prime \textrm{rgb}}$) and produce color-related feature $f^c$. Then our feature discriminator $D_F$ is developed to determine whether the input color-invariant features ($f^s_{\textrm{rgb}}$ and $f^s_{\textrm{gray}}$) are from same distribution. Finally, the generator $G$ jointly takes the color-invariant ($f^s_{\textrm{gray}}$) derived from gray-scaled image and color related feature ($f^c_{\textrm{rgb}}$) from RGB inputs, producing the synthesized RGB output images ${\hat{x}}_{\textrm{rgb}}$ while jointly training with additional image discriminator ($D_I$).}
  \label{fig:Model}
\end{figure*}

\section{CASE-Net}\label{sec:method}


For the sake of the completeness, we define the notations to be used in this paper. 
In the training stage, we have access to a set of~$N$ RGB images $X_{\textrm{rgb}} = \{x_i^{\textrm{rgb}}\}_{i=1}^N$ and its corresponding label set $Y_{\textrm{rgb}} = \{y_i^{\textrm{rgb}}\}_{i=1}^N$, where $x_i^{\textrm{rgb}} \in \bbR^{H \times W \times 3}$ and $y_i^{\textrm{rgb}} \in \mathbb{N}$ are the $i^\mathrm{th}$ RGB image and its label, respectively. To allow our model to handle images of different color variations, we generate a gray-scaled image set $X_{\textrm{gray}} = \{x_i^L\}_{i=1}^N$ by multiplying each image from $X_{\textrm{rgb}}$ with RGB channel summation factors, followed by duplicating the single channel back to the original image size (\ie $x_i^{\textrm{gray}} \in \bbR^{H \times W \times 3}$). Naturally, the label set $Y_{gray}$ for $X_{gray}$ is identical to $Y_{rgb}$. In order to achieve body-shape distilling via image generation, we also sample another set of RGB images $X^{\prime}_{rgb} = \{x_i^{\prime rgb}\}_{i=1}^N$, where its corresponding label set $Y^{\prime}_{rgb} = \{y_i^{\prime rgb}\}_{i=1}^N$ is same as $Y_{rgb}$ but with different pose and view point.

As depicted in Figure~\ref{fig:Model}, CASE-Net consists of five components:(1) the shape encoder $E_S$, (2) the color encoder $E_C$, (3) the feature discriminator $D_F$, (4) the image generator $G$, and (5) the image discriminator $D_I$. We now describe how these models work together to learn a body shape feature which can be used for re-ID in domains that do not use color. Training CASE-Net results in learning a shape encoding and a color encoding of an image of a person. However, we are primarily interested in the body shape feature since it can be re-used for cross-domain (non-color dependent) re-ID tasks.

\subsection{Clothing color adaptation in re-ID}

\paragraph{\textbf{Shape encoder ($E_S$).}}
To utilize labeled information of training data for person re-ID, we employ classification loss on the output feature vector $f^s$ ($f_{rgb}$ and $f_{gray}$). With person identity as ground truth information, we can compute the negative log-likelihood between the predicted label $\tilde{y} \in \bbR^K$ and the ground truth one-hot vector $\hat{y} \in \mathbb{N}^K$, and define the identity loss $\mathcal{L}_{id}$ as
\begin{equation}
  \begin{aligned}
  \mathcal{L}_{id}
  &~ =  - \mathbb{E}_{(x_{\textrm{rgb}},y_{\textrm{rgb}}) \sim (X_{\textrm{rgb}},Y_{\textrm{rgb}})}\sum_{k=1}^{K}\hat{y}_k^{\textrm{rgb}}\log(\tilde{y}_k^{\textrm{rgb}})\\
  &~ - \mathbb{E}_{(x_{\textrm{gray}},y_{\textrm{gray}}) \sim (X_{\textrm{gray}},Y_{\textrm{gray}})}\sum_{k=1}^{K}\hat{y}_k^{\textrm{gray}}\log(\tilde{y}_k^{\textrm{gray}}),
  \end{aligned}
  \label{eq:cls}
\end{equation}
where $K$ is the number of identities (classes).

To further enhance the discriminative property, we impose a triplet loss $\mathcal{L}_{tri}$ on the feature vector $f^s$, which would maximize the inter-class discrepancy while minimizing intra-class distinctness. To be more specific, for each input image $x$, we sample a positive image $x_\mathrm{pos}$ with the same identity label and a negative image $x_\mathrm{neg}$ with different identity labels to form a triplet tuple. Then, the following equations compute the distances between $x$ and $x_\mathrm{pos}$/$x_\mathrm{neg}$:
\begin{equation}
  \begin{aligned}
  d_\mathrm{pos} = \|{f}_x - {f}_{x_\mathrm{pos}}\|_2,
  \end{aligned}
  \label{eq:d-pos}
\end{equation}
\begin{equation}
  \begin{aligned}
  d_\mathrm{neg} = \|{f}_x - {f}_{x_\mathrm{neg}}\|_2,
  \end{aligned}
  \label{eq:d-neg}
\end{equation}
where ${f}_x$, ${f}_{x_\mathrm{pos}}$, and ${f}_{x_\mathrm{neg}}$ represent the feature vectors of images $x$, $x_\mathrm{pos}$, and $x_\mathrm{neg}$, respectively. With the above definitions, we have the triplet loss $\mathcal{L}_{tri}$ defined as
\begin{equation}
  \small
  \begin{aligned}
  \mathcal{L}_{tri} &~ 
  = \mathbb{E}_{(x_{\textrm{rgb}},y_{\textrm{rgb}}) \sim (X_{\textrm{rgb}},Y_{\textrm{rgb}})}\max(0, m + d_\mathrm{pos}^{\textrm{rgb}} - d_\mathrm{neg}^{\textrm{rgb}}) \\
  + &~ \mathbb{E}_{(x_{\textrm{gray}},y_{\textrm{gray}}) \sim (X_{\textrm{gray}},Y_{\textrm{gray}})}\max(0, m + d_\mathrm{pos}^{\textrm{gray}} - d_\mathrm{neg}^{\textrm{gray}}),
  \end{aligned}
  \label{eq:tri}
\end{equation}
where $m > 0$ is the margin used to define the distance difference between the distance of positive image pair $d_\mathrm{pos}$ and the distance of negative image pair $d_\mathrm{neg}$.

\paragraph{\textbf{Feature discriminator ($D_F$).}}
Next, since our goal is to derive body-shape representations which do not depend on clothing-color, we first learn color-invariant representation by encouraging the content encoder $E_S$ to generate similar feature distributions when observing both $X_{\textrm{rgb}}$ and $X_{\textrm{gray}}$. To achieve this, we advance adversarial learning strategies and deploy a feature discriminator ${D}_{F}$ in the latent \emph{feature space}. This discriminator ${D}_{F}$ takes the feature vectors $f_{\textrm{rgb}}$ and $f_{\textrm{gray}}$ as inputs to determine whether the input feature vectors are from $X_{\textrm{rgb}}$ or $X_{\textrm{gray}}$. To be more precise, we define the feature-level adversarial loss $\mathcal{L}_\mathrm{adv}^{{D}_{F}}$ as
\begin{equation}
  \begin{aligned}
  \mathcal{L}_\mathrm{adv}^{{D}_{F}} = &~ \mathbb{E}_{x_{\textrm{rgb}} \sim X_{\textrm{rgb}}}[\log({D}_{F}(f_{\textrm{rgb}}))]\\
  + &~ \mathbb{E}_{x_{\textrm{gray}} \sim X_{\textrm{gray}}}[\log(1 - {D}_{F}(f_{\textrm{gray}}))],
  \end{aligned}
  \label{eq:adv_loss_feature}
\end{equation}
where $f_{\textrm{rgb}}={E_S}({x_{\textrm{rgb}}})$ and $f_{\textrm{gray}}={E_S}({x_{\textrm{gray}}}) \in \bbR^{d}$ denote the encoded RGB and gray-scaled image features, respectively.\footnote{For simplicity, we omit the subscript $i$, denote RGB and gray-scaled images as $x_{\textrm{rgb}}$ and $x_{\textrm{gray}}$, and represent their corresponding labels as $y_{\textrm{rgb}}$ and $y_{\textrm{gray}}$.} With loss $\mathcal{L}_\mathrm{adv}^{{D}_{F}}$, our feature discriminator ${D}_{F}$ distinguish the features from two distributions while our shape encoder $E_S$ aligns the feature distributions across color variations, carrying out the learning of color-invariant representations for clothing via adversarial manner. 


\subsection{Pose guidance for body shape disentanglement}

\paragraph{\textbf{Color encoder ($E_C$).}}
To ensure our derived feature is body-shape related in clothing-color changing tasks, we need to perform additional body-shape disentanglement during the learning of our CASE-Net. That is, we have the color encoder $E_C$ in Fig.~\ref{fig:Model} encodes the inputs from RGB images set $X^{\prime}_{\textrm{rgb}}$ into color-related features ${f}_{\textrm{rgb}}^c$. As a result, both gray-scaled body-shape and color features would be produced in the latent space.
Inspired by DG-Net~\cite{zheng2019joint} using gray-scaled image to achieve body-shape disentanglement across pose variations, we similarly enforce the our generators $G$ to produce the person images conditioned on the encoded color feature coming from different pose. To be precise, we have the generator take the concatenated shape and color feature pair $(f_{\textrm{gray}}^s,f_{\textrm{rgb}}^c)$ and output the corresponding image $\hat{x}_{\textrm{rgb}}$. 

\paragraph{\textbf{Image generator ($G$).}}
Since we have ground truth labels (i.e., image pair correspondences) from the training data, we can perform a image recovery task given two images $x_{\textrm{rgb}}$ and ${x_{\textrm{rgb}}}^{\prime}$ of the same person but with different poses, we expect that they share the same body-shape feature $f^c_{\textrm{gray}}$. Given the desirable feature pair $(f_{\textrm{gray}}^s,f_{\textrm{rgb}}^c)$, we then enforce $G$ to output the image $\hat{x}_{\textrm{rgb}}$ using the body-shape feature $f_{\textrm{gray}}^s$ which is originally associated with $x_{\textrm{rgb}}$. This is referred to as \textit{Pose guided} image recovery.

With the above discussion, image reconstruction loss $\mathcal{L}_\mathrm{rec}$ can be calculated as:

\begin{equation}
  \label{eq:rec}
  \begin{aligned}
  \mathcal{L}_\mathrm{rec} = &~ \mathbb{E}_{x_{rgb} \sim X_{rgb}, x_{gray} \sim X_{gray}, x_{rgb}^{\prime} \sim X_{rgb}^{\prime}}[\|\hat{x}_{rgb} - x_{rgb}\|_1],
  \end{aligned}
\end{equation}
where $\hat{x}_{rgb}$ denotes $\hat{x}_{rgb} = G(f_{gray}^s,f_{rgb}^c)$. Note that we adopt the L$1$ norm in the above reconstruction loss terms as it preserves image sharpness~\cite{huang2018munit}. 

\paragraph{\textbf{Image discriminator ($D_I$).}}
To further enforce $G$ perform perceptual content recovery, we produce perceptually realistic outputs by having the image discriminator $D_I$ discriminate between the real images ${x}_{rgb}$ and the synthesized ones $\hat{x}_{rgb}$. To this end, we have both reconstruction loss and perceptual discriminator loss for image recovery. Thus, the image perceptual discriminator loss $\mathcal{L}_{adv}^{D_I}$ as
\begin{equation}
  \begin{aligned}
  \mathcal{L}_{adv}^{D_I} = &~ \mathbb{E}_{x_{rgb} \sim X_{rgb}}[\log({D}_{I}(x_{rgb}))]\\
  + & ~ \mathbb{E}_{x_{gray} \sim X_{gray}, x_{rgb}^{\prime} \sim X_{rgb}^{\prime}}[\log(1 - {D}_{I}(\hat{x}_{rgb}))].
  \end{aligned}
  \label{eq:adv_image_loss}
\end{equation}

To perform person re-ID in the testing phase, our network encodes the query image by $E_S$ for deriving the body shape feature ${f}_s$, which is applied for matching the gallery ones via nearest neighbor search (in Euclidean distances). We will detail the properties of each component in the following subsections.

It is import to note that the goal of CASE-Net is to perform re-ID in clothing changing scenario without observing ground true clothing changing training data. By introducing the aforementioned network module, our CASE-Net would be capable of performing re-ID in environments with clothing changes. More precisely, with the joint training of encoders/generator and the feature discriminator, our model allows learning of body-structural representation. The pseudo code for training our CASE-Net using above losses is summarized in Algorithm \ref{alg:pdanet}, where $\lambda_{tri}$ and $\lambda_{I}$ are hyper-parameters.
\begin{algorithm}[t]
\small
\KwData{Image set: $X_{rgb}$, $X_{gray}$, $X_{rgb}^{\prime}$; Label set: $Y_{rgb}$, $Y_{gray}$
}
\KwResult{Configurations of CASIE-Net}
$\theta_{E_S}$, $\theta_{E_C}$, $\theta_{D_F}$, $\theta_{G}$, $\theta_{D_I}$ $\leftarrow$ initialize\\
  \For{Num. of training Iters.}{
    $x_{rgb}$, $x_{gray}$, $x_{rgb}^{\prime}$, $y_{rgb}$, $y_{gray}$  $\leftarrow$ sample from $X_{rgb}$, $X_{gray}$, $X_{rgb}^{\prime}$, $Y_{rgb}$, $Y_{gray}$\\
    $f_{rgb}^s$, $f_{gray}^s$,$f_{rgb}^{c}$ $\leftarrow$ obtain by  $E_S(x_{rgb})$, $E_S(x_{gray})$, $E_C(x_{rgb}^{\prime})$\\
    %
    %
    $\mathcal{L}_\mathrm{id}$, $\mathcal{L}_{tri} \leftarrow$ calculate by (\ref{eq:cls}), (\ref{eq:tri})\\
    $\theta_{E_S} \xleftarrow{+} -{\nabla}_{\theta_{E_S}}(\mathcal{L}_\mathrm{id}+ \lambda_{tri} \mathcal{L}_{tri})$\\
    $\hat{x}_{rgb}$ $\leftarrow$ obtain by $G(f_{gray}^s,f_{rgb}^{c})$\\
    $\mathcal{L}_\mathrm{adv}^{D_F}$, $\mathcal{L}_{rec}$, $\mathcal{L}_\mathrm{adv}^{D_I}$ $\leftarrow$ calculate by  (\ref{eq:adv_loss_feature}), (\ref{eq:rec}), (\ref{eq:adv_image_loss})\\
    %
    \For{  Iters. of updating generator }{
    $\theta_{E_S} \xleftarrow{+} -{\nabla}_{\theta_{E_S}}(- \mathcal{L}_\mathrm{adv}^{D_F})$\\
    $\theta_{E_S, E_C, G} \xleftarrow{+} -{\nabla}_{\theta_{E_S, E_C, G}}(\mathcal{L}_{rec} - \lambda_{I}\mathcal{L}_\mathrm{adv}^{D_I})$\\
  	}
    \For{Iters. of updating discriminator }{
    $\theta_{D_F} \xleftarrow{+} -{\nabla}_{\theta_{D_F}}\mathcal{L}_\mathrm{adv}^{D_F}$\\
    $\theta_{D_I} \xleftarrow{+} -{\nabla}_{\theta_{D_I}}\mathcal{L}_\mathrm{adv}^{D_I}$\\
  	}
  }
\caption{Learning of CASIE-Net}\label{alg:pdanet}
\normalsize
\end{algorithm}



\begin{table}[t]
  \caption{\textbf{Quantitative results of person re-ID on the SMPL-reID and Div-Market dataset.} Note that all the reported results are reproduced using released codes available online. *indicates replacement of IR images with gray-scaled ones during training.}
  \centering
  \label{table:dg}
  \resizebox{\linewidth}{!}
  {
  \begin{tabular}{l|cccc|cccc}
  \toprule
  \multirow{2}{*}{Method} & \multicolumn{4}{c|}{SMPL-reID} & \multicolumn{4}{c}{Div-Market} \\
  %
  \cmidrule{2-9} 
  & R1 & R5 & R10 & mAP & R1 & R5 & R10 & mAP\\
  \midrule
  Verif-Identif~\cite{zheng2018discriminatively} (TOMM'17) & 19.0 & 35.6 & 43.9 & 4.2 & 9.2 & 23.9 & 34.6 & 1.0 \\
  SVDNet~\cite{sun2017svdnet} (ICCV'17) & 20.7 & 46.0 & 59.4 & 5.3 & 9.8 & 25.1 & 35.5 & 1.3 \\
  FD-GAN~\cite{ge2018fd} (NIPS'18) & 21.2 & 46.5 & 59.9 & 5.1 & 14.3 & 26.4 & 36.5 & 1.6\\
  Part-aligned~\cite{suh2018part} (ECCV'18) & 23.7 & 47.3 & 60.6 & 5.5 & 14.9 & 27.4 & 36.1 & 1.8\\
  PCB~\cite{sun2018beyond} (ECCV'18) & 25.5 & 48.9 & 61.9 & 5.9 & 15.7 & 27.0 & 39.5 & 1.7\\
  DG-Net~\cite{zheng2019joint} (CVPR'19) &  27.2 & 51.3 & 63.3 & 6.2  & 19.7 & 30.1 & 47.5 & 2.2\\
  \midrule
  cmGAN*~\cite{dai2018cross} (IJCAI'18) & 29.6 & 55.5 & 65.3 & 8.2 & 23.7 & 37.0 & 50.5 & 2.9\\
  AlignGAN*~\cite{wang2019rgb} (ICCV'19) & 41.8 & 63.3 & 72.2 & 12.6 & 26.0 & 45.3 & 61.0 & 3.4\\
  \midrule
  %
  %
  Ours & \textbf{62.0} & \textbf{77.8} & \textbf{81.5} & \textbf{28.1} & \textbf{56.2} & \textbf{71.5} & \textbf{79.2} & \textbf{13.5}\\
  \bottomrule
  \end{tabular}
  }
\end{table}
\begin{table*}[t]
  \caption{\textbf{Quantitative results of person re-ID on the Market1501 dataset.} \emph{Left block}: standard re-ID evaluation. \emph{Right block}: extended re-ID setting. Note that all the reported results are reproduced using released codes available online. *indicates replacement of IR images with gray-scaled ones during training. }
  \centering
  \label{table:market}
  \resizebox{\linewidth}{!}
  {
  \begin{tabular}{l|cccc|cccc|cccc|cccc}
  \toprule
  \multirow{3}{*}{Method} & \multicolumn{4}{c|}{Standard re-ID } & \multicolumn{12}{c}{Extended re-ID }\\\cmidrule{2-17} 
  & \multicolumn{4}{c|}{Q: RGB, G: RGB}& \multicolumn{4}{c|}{Q: Gray, G: RGB}& \multicolumn{4}{c|}{Q: RGB, G: Gray}& \multicolumn{4}{c}{Q: Gray, G: Gray}\\
  & R1 & R5 & R10 & mAP & R1 & R5 & R10 & mAP & R1 & R5 & R10 & mAP & R1 & R5 & R10 & mAP\\
  \midrule
  Verif-Identif~\cite{zheng2018discriminatively} (TOMM'17) & 79.5 & 86.0 &90.3 & 61.5 & 10.2 & 15.4 & 21.1 & 7.8 & 19.5 & 35.6 & 43.9 & 10.9 & 42.5 & 61.3 & 74.2 & 20.6\\
  SVDNet~\cite{sun2017svdnet} (ICCV'17) & 82.2 & 92.3 & 93.9 & 62.4 & 10.1 & 13.2 & 22.5 & 8.9 & 18.9 & 36.5 & 45.4 & 11.0 & 42.0 & 62.7 & 72.1 & 21.1\\
  FD-GAN~\cite{ge2018fd} (NIPS'18) & 90.5 & 96.0 & 97.7 & 77.9 & 12.4 & 19.6 & 23.8 & 10.1 & 30.5 & 50.1 & 59.6 & 18.4 & 49.7 & 69.8 & 76.2 & 23.2\\
  Part-aligned~\cite{suh2018part} (ECCV'18) & 93.8 & 97.7 & 98.3 & 79.9 & 14.1 & 22.5 & 27.9 & 11.6 & 36.6 & 58.7 & 67.4 & 20.0 & 51.3 & 73.4 & 80.4 & 26.5\\
  PCB~\cite{sun2018beyond} (ECCV'18) & 93.2 & 97.3 & 98.2 & 81.7 & 13.6 & 22.4 & 27.4 & 10.6 & 35.5 & 56.2 & 65.1 & 19.3 & 50.2 & 72.9 & 80.1 & 26.2\\
  DG-Net~\cite{zheng2019joint} (CVPR'19) & {94.4} & {98.4} & {98.9} & {85.2} & 15.1 & 23.6 & 29.4 & 12.1 & 37.7 & 59.8 & 68.5 & 22.9 & 52.9 & 73.8 & 81.5 & 27.5\\
  \midrule
  cmGAN*~\cite{dai2018cross} (IJCAI'18) & 82.1 & 92.5 & 94.1 & 61.8 & 67.2 & 83.5 & 88.6 & 46.3 & 70.4 & 86.8 & 91.5 & 46.5 & 70.8 & 86.2 & 90.1 & 46.7\\
  AlignGAN*~\cite{wang2019rgb} (ICCV'19) & 89.3 & 95.4 & 97.2 & 74.3 & 77.2 & 89.6 & 94.7 & 57.0 & 79.4 & 90.5 & 92.1 & 55.1 & 79.8 & 91.8 & 94.0 & 57.1\\
  \midrule
  %
  %
  Ours & \textbf{94.6} & \textbf{98.9} & \textbf{99.1} & \textbf{85.7} & \textbf{80.4} & \textbf{93.0} & \textbf{95.9} & \textbf{60.3} & \textbf{81.4} & \textbf{93.9} & \textbf{97.4} & \textbf{60.8} & \textbf{81.6} & \textbf{93.7} & \textbf{95.7} & \textbf{60.5}\\
  \bottomrule
  \end{tabular}
  }
\end{table*}
\section{Experiments}\label{sec:exps}

\subsection{Datasets}\label{sec:dataset}

To evaluate our proposed method, we conduct experiments on two of our synthesized datasets: SPML-reID and Div-Marke, and two benchmark re-ID datasets: Market-1501~\cite{zheng2015scalable}
and DukeMTMC-reID~\cite{zheng2017unlabeled,ristani2016MTMC}
, which is commonly considered in recent re-ID tasks. We also additionally conduct experiments on one cross-modality dataset named SYSU-MM01~\cite{wu2017rgb} to assess \emph{the generalization of our model when it learns body shape representation}.

\paragraph{\textbf{SPML-reID.}}
SPML-reID is our synthetic dataset to simulate clothing change for person re-ID. We render $100$ identity across $6$ view-points (different shooting angle from top view) and $10$ walking poses using SMPL\cite{Loper2015}.
Details of the SMPL model can be found at \cite{Loper2015}. 
For each identity, we render it using \cite{Kato2018} and pair it with a selected background image. For shape parameters, we sampled from the "walking" class from the AMASS dataset~\cite{Mahmood2019}. The identities are rendered from 6 different view points (see examples in Fig.~\ref{fig:smpl}), and $50$ identities are for training where no clothes change occurs, while the other $50$ identities in testing dataset contain clothes changes, totally $4764$ images.

\paragraph{\textbf{Div-Market.}} Div-Market is our small synthesized dataset from current Market-1501. We use our generative model similar as \cite{zheng2019joint} to change the clothing-color in the images of each identity. It contains total 24732 images of 200 identities each with hundreds of figures and \emph{it is only used for testing scenario}.

\paragraph{\textbf{Market-1501.}} The Market-1501~\cite{zheng2015scalable} is composed of 32,668 labeled images of 1,501 identities collected from 6 camera views. The dataset is split into two non-over-lapping fixed parts: 12,936 images from 751 identities for training and 19,732 images from 750 identities for testing. In testing, 3368 query images from 750 identities are used to retrieve the matching persons in the gallery. 

\paragraph{\textbf{DukeMTMC-reID.}} The DukeMTMC-reID~\cite{zheng2017unlabeled,ristani2016MTMC} is also a large-scale Re-ID dataset. It is collected from 8 cameras and contains 36,411 labeled images belonging to 1,404 identities. It also consists of 16,522 training images from 702 identities, 2,228 query images from the other 702 identities, and 17,661 gallery images. 

\paragraph{\textbf{SYSU-MM01.}} The SYSU-MM01~\cite{wu2017rgb} dataset is the first benchmark for cross-modality (RGB-IR) Re-ID, which is captured by 6 cameras, including two IR cameras and four RGB ones. This dataset contains 491 persons with total 287,628 RGB images
and 15,792 IR images from four RGB cameras and two IR
cameras. The training set consists of total 32,451 images
including 19,659 RGB images and 12,792 IR images, where the training set contains 395 identities and the test set contains 96 identities.

\subsection{Implementation Details}

We implement our model using PyTorch. Following Section~\ref{sec:method}, we use ResNet-$50$ pre-trained on ImageNet as our backbone of shape encoder $E_S$ and color encoder $E_C$. Given an input image $x$ (all images are resized to size $256 \times 128 \times 3$, denoting width, height, and channel respectively.), $E_S$ encodes the input into $2048$-dimension content feature $f^s$. The structure of the generator is $6$ convolution-residual blocks similar to that proposed by Miyato~\etal~\cite{miyato2018cgans}. The structure of the image discriminator ${D}_I$ employs the ResNet-$18$ as backbone while the architecture of shared feature discriminator $D_F$ adopts is composed of $5$ convolution blocks in our CASE-Net. All five components are all randomly initialized. The margin for the $\mathcal{L}_{tri}$ is set as $2.0$, and we fix $\lambda_{tri}$ and $\lambda_{I}$ as $1.0$ and $0.1$, respectively. The performance of our method can be possibly further improved by applying pre/post-processing methods, attention mechanisms, or re-ranking techniques. However, such techniques are not used in all of our experiments.

\subsection{Evaluation Settings and Protocol.} 

For our rendered \textbf{SPML-reID}, we train the model on training set and then inference it with testing set. For our synthesized testing set \textbf{Div-Market}, we evaluate the models training only with Market-1501 on the clothing-color changing dataset during the testing scenario. For \textbf{Market-1501}
, we augment the testing dataset by converting the RGB images into Gray-scaled ones. That is, in addition to the standard evaluation setting where both Probe (Query) and Gallery are of RGB, we conducted extended experiments on Gray/RGB, Gray/Gray, and Gray/Gray as Probe/Gallary sets for evaluating the generalization of current re-ID models. For \textbf{SYSU-MM01}, there are two test modes, \emph{i.e.}, all-search mode and indoor-search mode. For the all-search mode, all testing images are used. For the indoor-search mode, only indoor images from 1st, 2nd, 3rd, 6th cameras are used. The single-shot and multi-shot settings are adopted in both modes. Both modes use IR images as the probe set and RGB images as the gallery set.

We employ the standard metrics as in most person Re-ID literature, namely the cumulative matching curve (CMC) used for generating ranking accuracy, and the mean Average Precision (mAP). We report rank-1 accuracy and mean average precision (mAP) for evaluation on both datasets.

\begin{table*}[t]
  \scriptsize
  \caption{\textbf{Quantitative results of person re-ID on the DukeMTMC-reID dataset.} \emph{Left block}: standard re-ID evaluation. \emph{Right block}: extended re-ID setting. Note that all the reported results are reproduced using released codes available online. *indicates replacement of IR images with gray-scaled ones during training.}
  \centering
  \label{table:duke}
  \resizebox{\linewidth}{!}
  {
  \begin{tabular}{l|cccc|cccc|cccc|cccc}
  \toprule
  \multirow{3}{*}{Method} & \multicolumn{4}{c|}{Standard re-ID} & \multicolumn{12}{c}{Extended re-ID}\\\cmidrule{2-17} 
  & \multicolumn{4}{c|}{Q: RGB, G: RGB}& \multicolumn{4}{c|}{Q: Gray, G: RGB}& \multicolumn{4}{c|}{Q: RGB, G: Gray}& \multicolumn{4}{c}{Q: Gray, G: Gray}\\
  & R1 & R5 & R10 & mAP & R1 & R5 & R10 & mAP & R1 & R5 & R10 & mAP & R1 & R5 & R10 & mAP\\
  \midrule
  Verif-Identif~\cite{zheng2018discriminatively} (TOMM'17) & 68.7 & 81.5 & 84.2 & 49.8 & 8.9 & 15.4 & 20.3 & 6.2 & 16.1 & 33.9 & 43.5 & 8.1 & 35.4 & 52.3 & 60.0 & 17.4\\
  SVDNet~\cite{sun2017svdnet} (ICCV'17) & 76.5 & 87.1 & 90.4 & 57.0 & 9.1 & 16.3 & 20.5 & 6.9 & 16.5 & 36.4 & 45.8 & 9.6 & 37.0 & 52.8 & 60.9 & 17.5\\
  FD-GAN~\cite{ge2018fd} (NIPS'18) & 80.8 & 89.8 & 92.7 & 63.3 & 9.4 & 17.1 & 22.1 & 7.8 & 19.5 & 36.9 & 46.2 & 11.0 & 35.1 & 53.2 & 61.8 & 18.1\\
  Part-aligned~\cite{suh2018part} (ECCV'18) & 83.5 & 92.0 & 93.9 & 69.2 & 10.5 & 16.9 & 20.6 & 7.4 & 20.1 & 38.0 & 46.3 & 10.9 & 34.7 & 52.4 & 61.5 & 18.8\\
  PCB~\cite{sun2018beyond} (ECCV'18) & 82.9 & 91.1 & 93.6 & 67.1 & 9.1 & 17.4 & 21.4 & 6.6 & 19.4 & 37.7 & 46.4 & 10.8 & 34.2 & 52.6 & 60.9 & 18.4\\
  DG-Net~\cite{zheng2019joint} (CVPR'19) & {86.3} & {93.2} & {95.5} & {75.1} & 12.5 & 18.6 & 23.9 & 8.1 & 21.5 & 38.3 & 47.2 & 12.5 & 38.6 & 55.1 & 64.0 & 21.5\\
  \midrule
  cmGAN*~\cite{dai2018cross} (IJCAI'18) & 74.1 & 86.2 & 88.5 & 54.1 & 58.8 & 76.1 & 79.3 & 38.2 & 60.5 & 77.0 & 83.6 & 39.0 & 60.6 & 76.9 & 81.7 & 39.0\\
  AlignGAN*~\cite{wang2019rgb} (ICCV'19) & 80.1 & 87.6 & 90.5 & 58.1 & 63.8 & 79.4 & 82.8 & 47.1 & 62.5 & 80.3 & 84.8 & 44.1 & 63.8 & 80.1 & 84.2 & 42.5\\
  \midrule
  %
  Ours &  \textbf{86.4} & \textbf{93.7} & \textbf{95.6} & \textbf{75.4} & \textbf{65.8} & \textbf{80.7} & \textbf{85.5} & \textbf{50.5} & \textbf{66.9} & \textbf{83.9} & \textbf{87.9} & \textbf{50.2} & \textbf{65.3} & \textbf{84.8} & \textbf{87.3} & \textbf{49.5}\\
  \bottomrule
  \end{tabular}
  }
  \vspace{-3mm}
\end{table*}

\subsection{Comparisons with State-of-the-art Approaches}
\begin{table*}[t]
  \scriptsize
  \caption{\textbf{Quantitative results of person re-ID on the cross-modality SYSU-MM01 dataset.} To evaluate the generalization of our model in addressing cross-modality re-ID, we compare with existing models working on IR-RGB scenario. *indicates the results are reproduced using the released codes. Bold and underlined numbers indicate top two results, respectively.}
  \centering
  \label{table:sysu}
  \resizebox{\linewidth}{!}
  {
  \begin{tabular}{l|cccc|cccc|cccc|cccc}
  \toprule
  \multirow{3}{*}{Method} & \multicolumn{8}{c|}{All-search} & \multicolumn{8}{c}{Indoor-search}\\\cmidrule{2-17} 
  & \multicolumn{4}{c|}{Single-shot}& \multicolumn{4}{c|}{Multi-shot}& \multicolumn{4}{c|}{Single-shot}& \multicolumn{4}{c}{Multi-shot}\\
  & R1 & R10 & R20 & mAP & R1 & R10 & R20 & mAP& R1 & R10 & R20 & mAP& R1 & R10 & R20 & mAP\\
  \midrule
  HOG~\cite{dalal2005histograms} (CVPR'05) & 2.8 & 18.3 & 32.0 & 4.2 & 3.8 & 22.8 & 37.6 & 2.2 & 3.2 & 24.7 & 44.5 & 7.3 & 4.8 & 29.1 & 49.4 & 3.5\\
  LOMO~\cite{liao2015person} (CVPR'15) & 3.6 & 23.2 & 37.3 & 4.5 & 4.7 & 28.3 & 43.1 & 2.3 & 5.8 & 34.4 & 54.9 & 10.2 & 7.4 & 40.4 & 60.4 & 5.6\\
  Two Stream Net~\cite{wu2017rgb} (ICCV'17) & 11.7 & 48.0 & 65.5 & 12.9 & 16.4 & 58.4 & 74.5 & 8.0 & 15.6 & 61.2 & 81.1 & 21.5 & 22.5 & 72.3 & 88.7 & 14.0\\
  One Stream Net~\cite{wu2017rgb} (ICCV'17) & 12.1 & 49.7 & 66.8 & 13.7 & 16.3 & 58.2 & 75.1 & 8.6 & 17.0 & 63.6 & 82.1 & 23.0 & 22.7 & 71.8 & 87.9 & 15.1\\
  Zero Padding~\cite{wu2017rgb} (ICCV'17) & 14.8 & 52.2 & 71.4 & 16.0 & 19.2 & 61.4 & 78.5 & 10.9 & 20.6 & 68.4 & 85.8 & 27.0 & 24.5 & 75.9 & 91.4 & 18.7\\
  cmGAN~\cite{dai2018cross} (IJCAI'18) & 27.0 & 67.5 & 80.6 & 27.8 & 31.5 & 72.7 & 85.0 & 22.3 & 31.7 & 77.2 & 89.2 & 42.2 & 37.0 & 80.9 & 92.3 & 32.8\\
  AlignGAN~\cite{wang2019rgb} (ICCV'19) & \underline{42.4} & \underline{85.0} & \underline{93.7} & \underline{40.7} & \underline{51.5} & \underline{89.4} & \underline{95.7} & \underline{33.9} & \textbf{45.9} & \textbf{87.6} & \textbf{94.4} & \textbf{54.3} & \textbf{57.1} & \textbf{92.1} & \textbf{97.4} & \textbf{45.3}\\
  \midrule
  %
  Ours & \textbf{42.9} & \textbf{85.7} & \textbf{94.0} & \textbf{41.5} & \textbf{52.2} & \textbf{90.3} & \textbf{96.1} & \textbf{34.5} & \underline{44.1} & \underline{87.3} & \underline{93.7} & \underline{53.2} & \underline{55.0} & \underline{90.6} & \underline{96.8} & \underline{43.4}\\
  \bottomrule
  \end{tabular}
  }
\end{table*}
\begin{table}[t]
  \begin{center}
  \caption{\textbf{Ablation study of the loss functions on the Div-Market dataset.} We note that, each row indicates the model with only one loss excluded.}
  %
  \label{table:exp-abla}
  \begin{tabular}{l|cccc}
  \toprule
  Method & Rank 1 & Rank 5 & Rank 10 & mAP\\
  \midrule
  Ours (full model) & \textbf{56.2} & \textbf{61.5} & \textbf{69.2} & \textbf{13.5}\\
  Ours w/o $\mathcal{L}_\mathrm{adv}^{D_I}$ & 55.7  & 60.4 & 66.8 & 10.1 \\
  Ours w/o $\mathcal{L}_{tri}$ & 54.0 & 58.1 & 66.3&  8.7 \\
  Ours w/o $\mathcal{L}_{id}$ & 50.5 & 57.6 & 65.5 &  8.2 \\
  Ours w/o $\mathcal{L}_\mathrm{adv}^{D_F}$ & 49.8 & 51.5 & 64.1 & 7.3 \\
  Ours w/o $\mathcal{L}_{rec}$ & 46.5 & 50.1 & 61.5 & 6.9 \\

  \bottomrule
  \end{tabular}
  \end{center}
  \vspace{-9.0mm}
\end{table}
\begin{figure*}[t]
  \centering
  \begin{subfigure}[b]{0.24\linewidth}
    \centering\includegraphics[width=\linewidth]{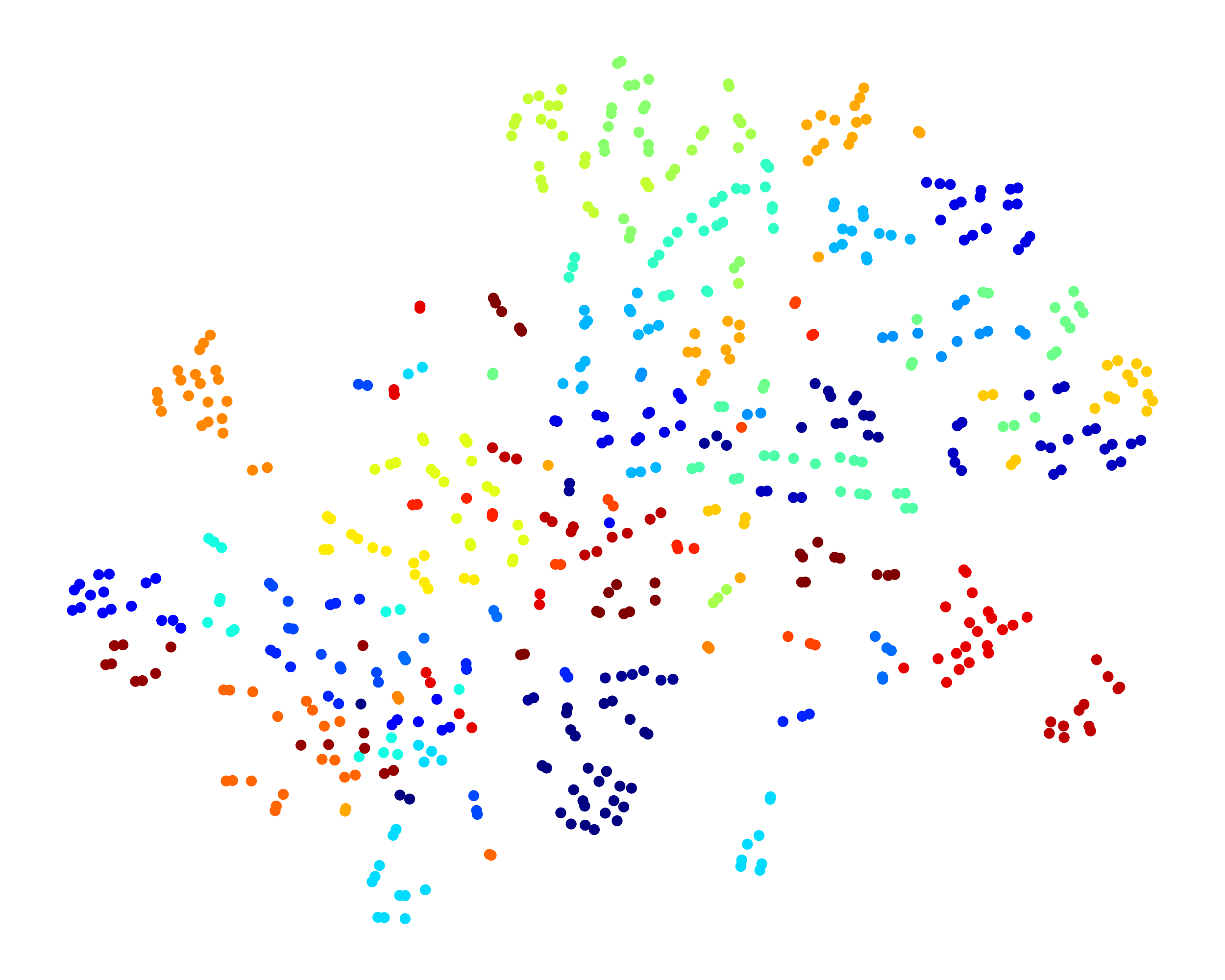}\\
    \caption{\cite{zheng2019joint} w.r.t. \textbf{identity}.}
    \label{fig:tsne-identity-dg}
  \end{subfigure}
  \begin{subfigure}[b]{0.24\linewidth}
    \centering\includegraphics[width=\linewidth]{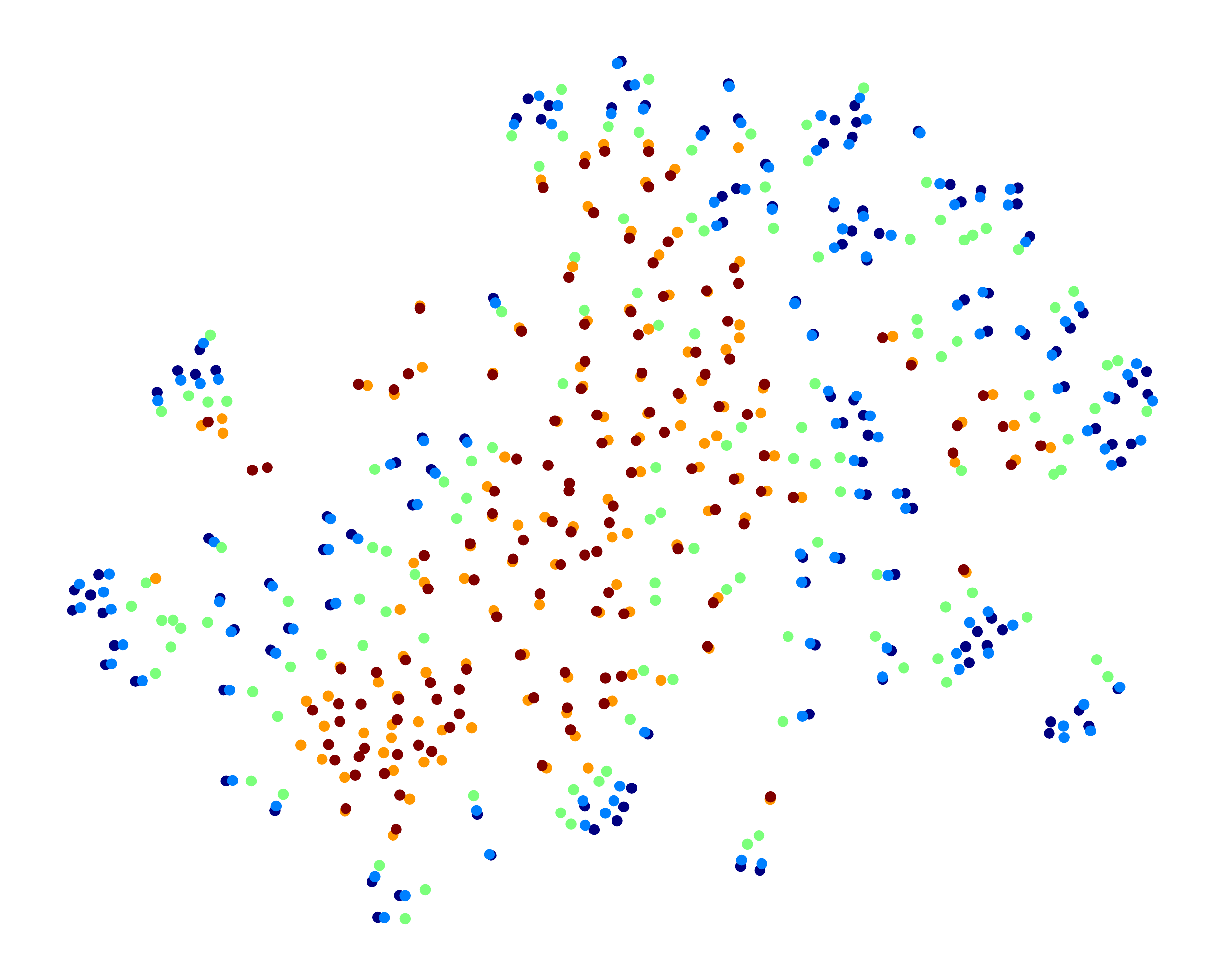}\\
    \caption{\cite{zheng2019joint} w.r.t. \textbf{clothing}.}
    \label{fig:tsne-color-dg}
  \end{subfigure}
  \begin{subfigure}[b]{0.24\linewidth}
    \centering\includegraphics[width=\linewidth]{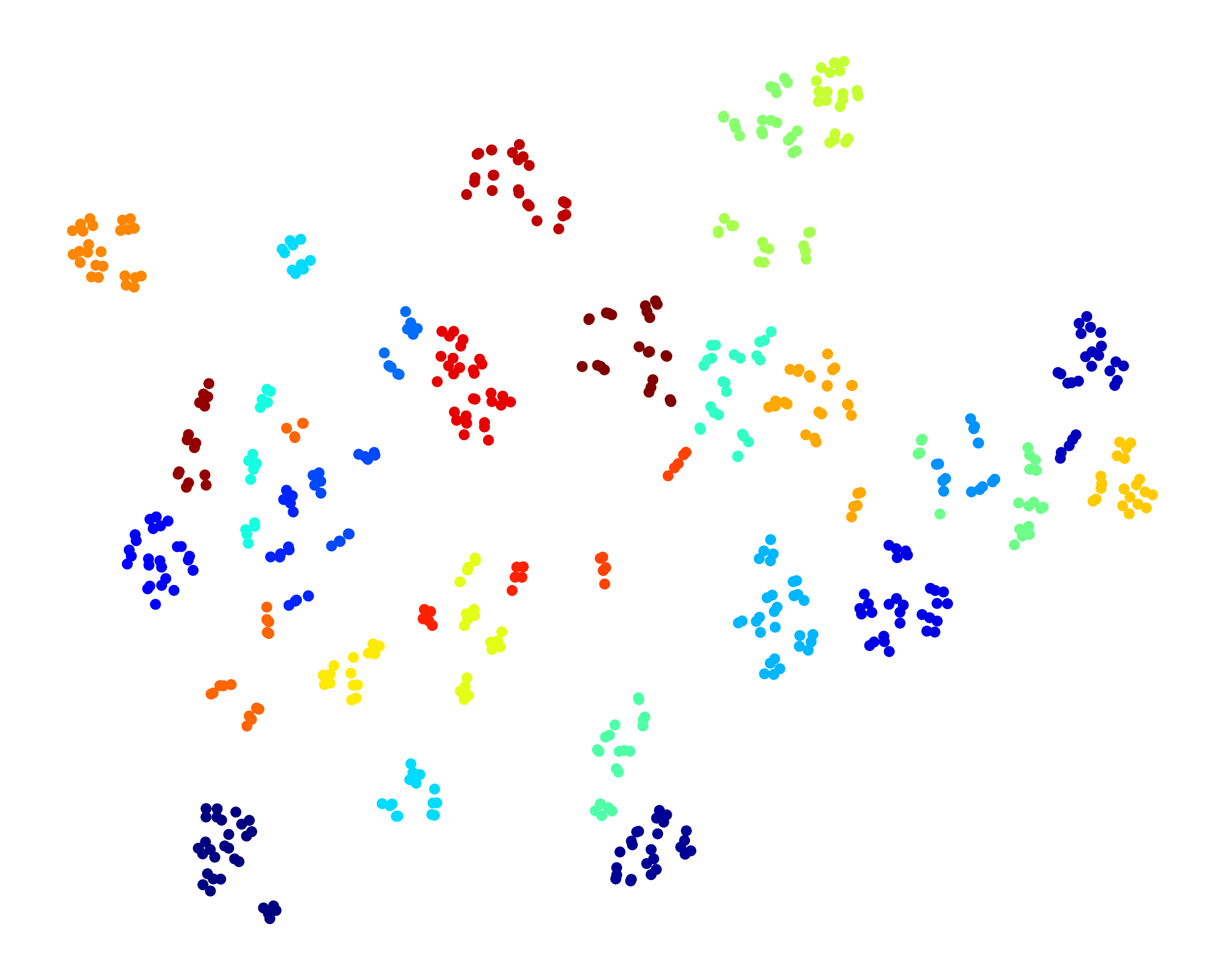}\\
    \caption{Ours w.r.t. \textbf{identity}.}
    \label{fig:tsne-identity-ours}
  \end{subfigure}
  \begin{subfigure}[b]{0.24\linewidth}
    \centering\includegraphics[width=\linewidth]{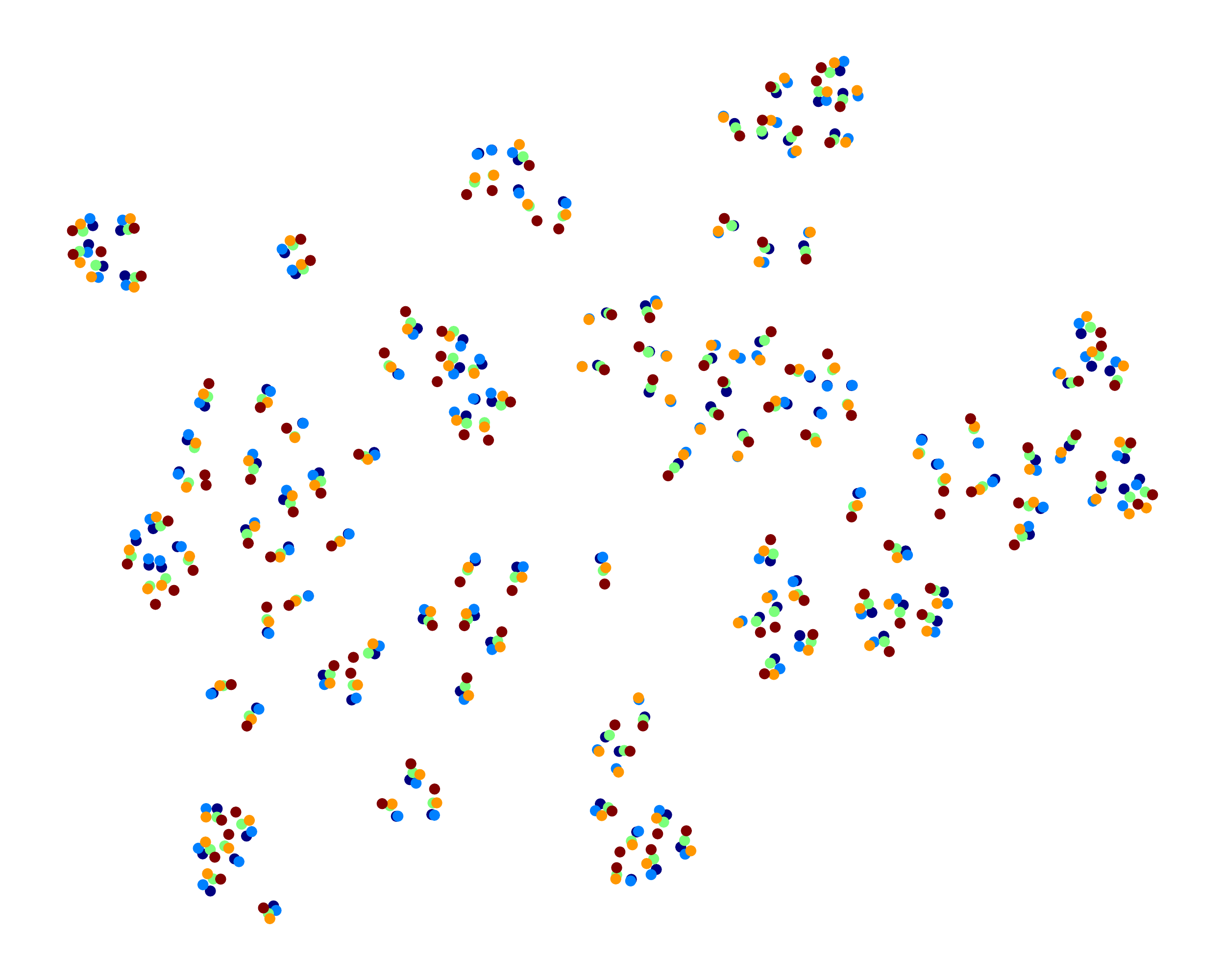}\\
    \caption{Ours w.r.t. \textbf{clothing}.}
    \label{fig:tsne-color-ours}
  \end{subfigure}
  \caption{Visualization of structure feature vectors $f^{s}$ on Div-Market via t-SNE.
  (a) $30$ different identities, each of which is shown in a unique color.
  (b) With five different appearance (clothing color) are considered and shown, images with the same dressing are shown in the same color.
  }
  \label{fig:tsne}
  \vspace{-5.0mm}
\end{figure*}

\paragraph{\textbf{SMPL-reID.}} To simulate the real-world clothing-color changing environment, we conducted the re-ID experiments on our SMPL-reID, and compared with the six state-of-the-art re-ID approaches and two cross-modality re-ID models. As the reported results presented on the left side of Table~\ref{table:dg}, our proposed CASE-Net outperforms all the compared methods by a large margin. In addition, some phenomenons can also be observed. First, we found severe performance drops in all the standard re-ID approaches, which indicates standard re-ID approaches all suffer from \emph{clothing-color/clothes} mismatch problems. Second, though two cross-modality methods demonstrate improvement, their models can not handle clothing-color changing in single modality either.

\paragraph{\textbf{Div-Market.}} For our synthesized Div-Market, we also compare our proposed method with six current standard re-ID approaches and two cross-modality re-ID models. We also reported the results on the right side of Table~\ref{table:dg}. Same phenomenons are also observed as SMPL-reID.

\paragraph{\textbf{Market-1501.}} We compare our proposed method with six current standard re-ID approaches and two cross-modality re-iD models whose codes are available online, and reported the results in one standard and three extended settings on the Market-1501. These standard approaches include Verif-Identif~\cite{zheng2018discriminatively}, SVDNet~\cite{sun2017svdnet}, Part-aligned~\cite{suh2018part}, FD-GAN~\cite{ge2018fd}, PCB~\cite{sun2018beyond}, and DG-Net~\cite{zheng2019joint} while cross-modality models involve cmGAN~\cite{dai2018cross} and AlighnGAN~\cite{wang2019rgb}. We report all the results in Table~\ref{table:market} and several phenomenons can be observed which we summarized as three folds. Firstly, state-of-the-arts methods outperform two cross-modality approaches by a margin but suffer severe performance drop in the extended evaluation, which shows their vulnerability to color variations and weak generalization when \emph{they train to overfit on the clothing color.} Second, our proposed CASE-Net outperforms all the methods in each settings, which demonstrates that its ability to derive body shape representation. 

\paragraph{\textbf{DukeMTMC-reID.}} For the DukeMTMC-reID dataset, we also compare our proposed method with six current standard re-ID approaches and two cross-modality re-ID models whose codes are available online, and we reported the results in one standard and three extended settings in as well Table~\ref{table:duke}. Same phenomenons are also observed as Market-1501.

\paragraph{\textbf{SYSU-MM01.}} To assess the generalization of our CASE-Net in cross-modality person re-ID, we also conducted additional experiments on the SYSU-MM01 dataset. We compare our proposed CASEI-Net with two hand-crafted features (HOG~\cite{dalal2005histograms}, LOMO~\cite{liao2015person}) and three cross-modality approaches (SYSU model~\cite{wu2017rgb}, cmGAN~\cite{dai2018cross},  AlighnGAN~\cite{wang2019rgb}). We reported the results in Table~\ref{table:sysu} and observe that our method achieves comparable result in the cross-modality re-ID setting. It to worth repeating that, our proposed CASE-Net which is developed for clothing-color changes in re-ID generalizes well in cross-modality re-ID.

\subsection{Ablation Studies}
\paragraph{\textbf{Loss functions.}}

To further analyze the importance of each introduced loss function, we conduct an ablation study shown in Table~\ref{table:exp-abla}. Firstly, the feature adversarial loss $\mathcal{L}_{rec}$ is shown to be vital to our CASE-Net, since we observe $20\%$ drops on Div-Market when the loss was excluded. This is caused by no explicit supervision to guide our CASE-Net to generate human-perceivable images with body shape disentanglement, and thus the resulting model would suffer from image-level information loss. Secondly, without the feature adversarial loss $\mathcal{L}_\mathrm{adv}^{D_F}$, our model would not be able to perform feature-level color adaptation, causing failure on learning clothing color invariant representation and resulting in the re-ID performance drop (about $15\%$). Thirdly, when either $\mathcal{L}_{id}$ or $\mathcal{L}_{tri}$ is turned off, our model is not able to be supervised using two re-ID losses, indicating that jointly use of two streams of supervision achieve best results. Lastly, the image adversarial loss $\mathcal{L}_\mathrm{adv}^{D_I}$ is introduced to our CASE-Net to mitigate the perceptual image-level information loss.

\paragraph{\textbf{Visualization.}}
We now visualize the feature vectors $f^s$ on our Div-Market in Figures~\ref{fig:tsne} via t-SNE. It is worth to repeat that, in our synthesized Div-Market same identity can have different wearings while several identities can have the same wearing. In the figure, we select $30$ different person identities, each of which is indicated by a color. From Fig.~\ref{fig:tsne-identity-dg} and Fig.~\ref{fig:tsne-identity-ours}, we observe that our projected feature vectors are well separated when it compared with DG-Net~\cite{zheng2019joint}, which suggests that sufficient re-ID ability can be exhibited by our model. On the other hand, for Fig.~\ref{fig:tsne-color-dg} and Fig.~\ref{fig:tsne-color-ours}, we colorize each same cloth dressing with a color. It can be observed that our projected feature vectors of the same identity but different dressing are all well clustered while the ones of DG-Net~\cite{zheng2019joint} are not.

\section{Conclusions}\label{sec:con}

In this paper, we have unfolded an challenge yet significant person re-identification task which has been long ignored in the past. We collect two re-ID datsets (\emph{SMPL-reID} and \emph{Div-Market}) for simulating real-world scenario, which contain changes in clothes or clothing-color. To address clothing changes in re-ID, we presented a novel Color Agnostic Shape Extraction Network (CASE-Net) which learns body shape representation training or fine-tuning on data containing clothing change. By advancing the adversarial learning and body shape disentanglement, our model resulted in satisfactory performance on the collected datasets (SPML-reID and Div-Market) and two re-ID benchmarks. Qualitative results also confirmed that our model is capable of learning body shape representation, which is clothing-color invariant. Furthermore, the extensive experimental result on one cross-modality dataset also demonstrated the generalization of our model to cross-modality re-ID.

{\small
\bibliographystyle{ieee_fullname}
\bibliography{main}
}

\end{document}